\documentclass[preprint,12pt]{elsarticle}
\usepackage{hyperref}
\usepackage{graphicx}
\usepackage{amssymb}
\usepackage{makecell}
\usepackage{multirow}
\usepackage{enumitem}
\newcommand{\etal}{\emph{et al. }}

\journal{Pattern Recognition(Accepted)}

\begin{document}

\begin{frontmatter}


\title{A Deep One-Shot Network for Query-based Logo Retrieval}

\author{Ayan Kumar Bhunia\textsuperscript{1} \hspace{.1cm} Ankan Kumar Bhunia\textsuperscript{2} \hspace{.1cm} Shuvozit Ghose\textsuperscript{3} \hspace{.1cm} Abhirup Das\textsuperscript{4} \hspace{.1cm} Partha Pratim Roy\textsuperscript{5} \hspace{.1cm} Umapada Pal\textsuperscript{6}}

\address{\textsuperscript{1}University of Surrey, England, United Kingdom. \hspace{.1cm} \textsuperscript{2} Jadavpur University, India 
\hspace{.1cm}  \\ \textsuperscript{3,4} Institute of Engineering \& Management, India  \hspace{.1cm} \textsuperscript{5} Indian Institute of Technology Roorkee, India \textsuperscript{6} Indian Statistical Institute, India}

\begin{abstract}
Logo detection in real-world scene images is an important problem with applications in advertisement and marketing. Existing general-purpose object detection methods require large training data with annotations for every logo class. These methods do not satisfy the incremental demand of logo classes necessary for practical deployment since it is practically impossible to have such annotated data for new unseen logo. In this work, we develop an easy-to-implement query-based logo detection and localization system by employing a one-shot learning technique using off the shelf neural network components. Given an image of a query logo, our model searches for logo  within a given target image and predicts the possible location of the logo by estimating a binary segmentation mask. The proposed model consists of a conditional branch and a segmentation branch. The former gives a conditional latent representation of the given query logo which is combined with feature maps of the segmentation branch at multiple scales in order to obtain the matching location of the query logo in a target image. Feature matching between the latent query representation and multi-scale feature maps of segmentation branch using simple concatenation operation followed by $1\times1$ convolution layer makes our model scale-invariant. Despite its simplicity, our query-based logo retrieval framework achieved superior performance in FlickrLogos-32 and TopLogos-10 dataset over different existing baseline methods. 
\end{abstract}


\end{frontmatter}


\section{Introduction}
\label{S:1}

Detection of logos in scene images and videos has a number of useful applications: commercial analysis of brands \cite{gao2014brand}, vehicle-logo detection \cite{psyllos2010vehicle} for intelligent traffic-control systems and even Augmented Reality \cite{hagbi2009shape}. A logo is a unique symbol representing any brand or organization that expresses its functionality and distinguishes the brand and its products from others. A merchant can easily assess the presence of his brand in television, social media and e-commerce sites by searching for the company's unique logo. Detection and localization of logos is a crucial step in inspecting market trends, allowing companies to meet the customers' needs by optimizing existing marketing schemes. 
Automating logo detection will also allow merchants to detect copyright infringement cases by testing the originality of suspicious advertisements.

\setlength\textfloatsep{5mm}
\begin{figure}
\label{pic1}
  \includegraphics[width=1\linewidth]{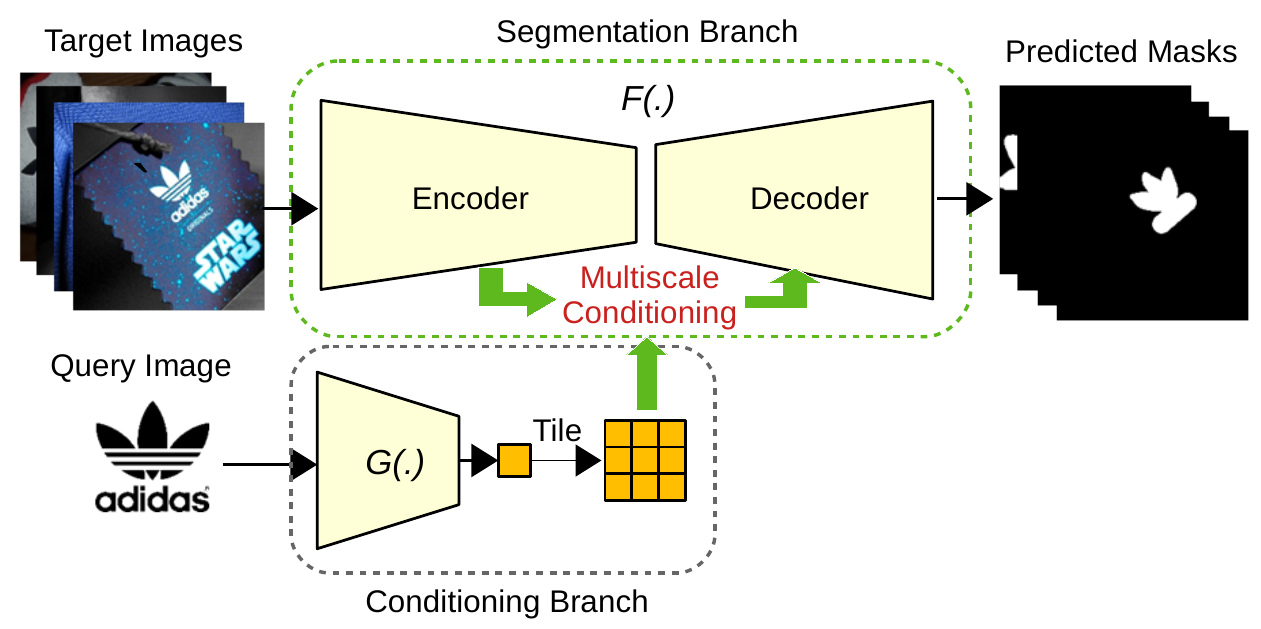}
  \caption{Illustration of Query-based logo detection problem.}
  \label{figure:1}

\end{figure}

Logo detection problem can be seen as a special case of object detection \cite{ren2015faster} in images. 
The appearance of a logo varies drastically in real-world images due to lighting effects, occlusions, rotations, shearing effects and scales. It is hard to detect different sizes of logos in a diverse contextual environment with uncontrolled illumination, low resolution, and high background clutter. In the recent years, logo detection has gained a lot of attention from the computer vision community \cite{bianco2017deep,boia2016logo,eggert2015benefit,oliveira2016automatic,su2017deep,alaei2016logo}. Earlier works of logo detection rely on the bag-of-words approach \cite{csurka2004visual} where SIFT features are quantized into a vocabulary of learned logo patterns in images. Boia \etal \cite{boia2016logo} used a novel approach based on homographic class graphs to perform both logo localization and recognition. Recently, significant improvement has been achieved by adopting deep-learning techniques in this field \cite{bianco2015logo,eggert2015benefit,iandola2015deeplogo}. 
In the meantime, several object detection algorithms have been introduced, namely, R-CNN \cite{girshick2014rich}, Fast R-CNN \cite{girshick2015fast} and Faster R-CNN \cite{ren2015faster} which have been successfully adapted for the logo detection problem \cite{oliveira2016automatic}, boosting object-recognition performances further. In addition, due to advent of deep learning, there have been significant progress in scene-text detection \cite{liu2019curved}, scene text recognition \cite{luo2019moran}, script identification \cite{bhunia2019script} tasks etc; however, there are very limited recent deep-frameworks towards logo detection and recognition in spite of its complexity.

Deep-learning based frameworks are largely data-driven, contrary to logo-datasets that have several image classes but few images. Since deployable logo detection models need to be robust to new unseen logos,  the model should be designed to satisfy the incremental demands for logo classes, contrary to existing methods which are limited to a set of seen logos and are not scalable enough for practical deployment. With the current problem setup, it is impossible for models to work with the logo of any brand. To meet the need for a scalable solution, we re-design the problem statement as shown in Figure \ref{pic1}. Given a query logo of a particular company or organization, the main objective would be to find out whether the logo of the same company or organization is present in a target image or not. If present, the model fetches its position within the given scene. 

Following this problem statement, we propose a one-shot learning based technique to design a framework that adapts to new logo classes in a data-efficient way. Recently, one-shot learning \cite{fei2006one} has gained notable attention in learning new concepts from sparse data.  
One-shot image classification \cite{koch2015siamese} and one-shot image segmentation \cite{caelles2017one,Dong2018,shaban2017one, Zhang2018} are some practical applications of it. Inspired by these works, we propose a one-shot query based logo detection framework that consists of two modules: the conditioning module and the segmentation module. The conditioning part takes a query logo image and obtains a latent representation which will be used as a conditional input to the segmentation branch. The segmentation network is a basic encoder-decoder architecture. In order to capture the multi-scale correlation between the query image and different regions of the target image, we concatenate the conditional latent representation with multiple layers of the encoder part(segmentation network), followed by $1\times1$ convolution. 
The obtained representation is further combined with the respective layers of the decoder part in order to better guide the decoder part of segmentation network to generate binary segmentation mask, forming a skip-connection like architecture \cite{ronneberger2015u}. When the model detects high similarity between a query and some particular region of the target image, the network tries to produce high response at the corresponding portion of the segmentation map. To apply our model for a new logo-class, unlike the fine-tuning approach \cite{caelles2017one} which may require large number of training samples with corresponding ground truth data, our approach needs only one sample of the logo of a concerned company or organisation. This query logo sample can be obtained very easily from the official logo design image or just by cropping an appropriate logo portion from a scene logo image. In this paper, we make following novel contributions: 

\begin{itemize}[leftmargin=*]
\item We propose a scalable solution for the logo detection problem by re-designing the traditional problem setting. We present a query-based logo search and detection system by employing a simple, fully differentiable one-shot learning framework which can be used for new logo classes without further training the whole network. To the best of our knowledge, ours is the first work to address one-shot query based deep learning framework for logo retrieval which is novel in the literature of logo-retrieval research. 

\item To deal with the logos of varying sizes, we propose a novel one-shot framework through \textit{multi-scale conditioning} that is specially designed to learn the similarity between the query image and target image at multiple scales and resolutions.
\end{itemize}


\section{Related Works}
\paragraph{Traditional methods:} Logo-related research has been carried out for over two decades in the area of computer vision and pattern recognition. Earlier, logo recognition and detection were categorized as a specific problem of object detection. The primary models were developed on geometric object features \cite{lamdan1988object} which relied on properties of objects such as lines, vertices, curves and shapes. Later on, properties of pixel value (luminance or color) were introduced as photometric object \cite{schmid1997local,gevers1999color,lowe2004distinctive} features, which eventually replaced the previous ones. These features were computed locally, could solve the problem of occlusion to an extent and were able to distinguish similar objects better way \cite{schmid1997local}. Thereafter many logo-related works were carried out with the methods of content-based indexing and retrieval in trademark databases. The main goal is to assist in trademark infringement detection by checking a newly designed trademark with registered logos in archives \cite{kim1998content,yin2002content,eakins2003shape, van2007layout}. The task of trademark recognition in videos is inherently harder due to loss of quality
of original logos during processing (e.g. color sub-sampling, video interlacing, motion blur, etc.). However in this case, it is assumed that the acquired images are of good quality and moderate distortion free. Kovar \etal \cite{kovar2002logo} applied a heuristic technique to discard sparse or small populated edge regions of the images and analysed the set of significant edges during logo detection. The work in \cite{den2003logo} deals with the logos that appear on the rigid planar surfaces having homogeneous colored background in images using Hough Transformation. Color histogram back projection is applied on candidate logo regions \cite{hall2004brand} to recognize candidate logos. Multidimensional receptive field histograms are also used to perform the task of logo recognition. Here, the most likely logo region is computed for every candidate region. Therefore, if a region in the image does not contain a logo, the identification precision gets reduced.

The traditional logo recognition models are well established on key-point based detectors and descriptors (specially SIFT). SIFT-based models \cite{lowe2004distinctive} basically take an image and transform it into a large collection of feature vectors that are invariant to affine transformations and even robust to different lighting conditions. One of its main characteristics is the ability to detect stable salient points in the image across multiple scales. On the basis of this, Lamberto \etal \cite{bagdanov2007trademark} proposed a representation of trademarks and video frame contents using SIFT feature-points mainly targeting to detect, localize and retrieve trademarks in a robust manner, irrespective of irregularities. The classification of retrieved trademarks is analyzed by matching a set of SIFT feature descriptors for each trademark instance with the features detected in every frame of the video. Kleban \etal \cite{kleban2008spatial} proposed a logo detection model by clustering matching spatial configurations of frequent local features and introducing spatial pyramid mining.  Alexis and Olivier presented a new content-based retrieval framework \cite{joly2009logo} using a thresholding strategy in order to improve the accuracy of query images. In \cite{meng2010interactive}, the author described the logo detection problem as a small object detection problem and solved it by interactive visual object search through mutual information maximization. Romberg and Lienhart \cite{romberg2013bundle} exploited a large scale recognition approach using feature bundling. As feature bundles carry more information about the image content than single visual words, they aggregated individual local features into bundles and then detected logos by querying the database of reference images based on features. Based on the analysis of the local features and basic structure, such as edges, curves, triangles, etc. Romberg \etal  \cite{romberg2011scalable} presented a system by  encoding and indexing the spatial layout of local features found in logos. Revaud \etal \cite{revaud2012correlation} introduced dedicated correlation-based burstiness model using a down-weight technique for noisy logo detections. Boia \etal \cite{boia2015elliptical, boia2016logo} smartly exploited homographic class graphs to analyze logo localization and recognition tasks. It is noteworthy that they used inverted secondary models to control inverted colors instances. Marcal \etal \cite{rusinol2010efficient} presents a robust queried-by-example logo retrieval system where logos are compactly described as a variant of the shape context descriptor. They perform k-NN search in the locality sensitive hashing database to retrieve logos. Jianlong \etal \cite{fu2010effective} exploits local features to form a visual codebook and build an inverted file to accelerate the indexing process. Then several groups are proposed according to the local feature type namely “point type”, “shape type” and “patch type”. Finally, adaptive feature selection with weight updating mechanism is used to perform logo retrieval. Jinqiao \etal \cite{wang2006logo} used k-means clustering to develop a visual logo dictionary, and next, latent semantic indexing and analysis is used for logo retrieval. Soysal \etal \cite{soysal2011joint} considered spatial similarity of local patterns by utilizing a descriptor for scene logo retrieval. 
 
\paragraph{Deep learning-based methods:}
Bianco \etal \cite{bianco2015logo} applied an unsupervised segmentation algorithm to produce a number of object proposals that are more likely to contain a logo object. Thereafter these object proposals are processed through a query expansion step in order to deal with the variation of logo instances. Finally, pre-trained CNN model with SVM classifier was used for logo recognition. Based on the previous Deep-CNN based pipeline, Eggert \etal \cite{eggert2015benefit} trained an SVM classifier on synthetic training examples to compare with a trained classifier on real images. The main objective of this approach was to demonstrate the benefit of using synthetic images during training. Iandola \etal \cite{iandola2015deeplogo} investigated several variations of GoogLeNet architecture including GoogLeNet with global average pooling and auxiliary classifier after each inception layers. Oliveira \etal \cite{oliveira2016automatic} exploited Fast R-CNN to detect graphic logos. To tackle the limitation of large-scale graphic logo datasets, transfer learning was used to leverage image representation. 

Very recently, Bianco \etal \cite{bianco2017deep} proposed a deep learning pipeline investigating the benefits of different training choices such as class-balancing, data augmentation, contrast normalization, sample-weighting and explicit modelling of background class. Su \etal \cite{su2017deep} has introduced a  framework to generate new training data for logo detection by synthesizing context and thus it intends to increase the robustness against unseen cluttered background. Based on the similar hypothesis, a more advanced framework \cite{su2018open} using Generative Adversarial Networks has been designed to generate context consistent logo images for training.  \cite{su2018scalable} proposed a incremental learning approach which discovers informative training images from noisy web data in order to improve the performance. In contrast to these recent works, we here intend to explore the existing benchmark logo-datasets and design a new one-shot deep framework for query-based logo retrieval.  

\paragraph{One shot learning:} Creditably, one-shot learning requires only a single annotated image to learn a new class. For few-shot learning, only a few examples of a class are needed to generalize knowledge for recognition. In recent years, one-shot learning is applied to various fields of computer vision such as image classification and visual question answering. 
The Siamese network architecture by Koch \etal \cite{koch2015siamese} has shown that few-shot image classification can outperform several classification baselines for a binary verification task. The siamese network makes use of two
shared network to extract features from two input images and a similarity score between the two feature representations decides the correspondence between the input images. Annother important recent work in few-shot classification is Matching networks \cite{vinyals2016matching} that learns to determine the correct class label for a given query image from unseen catagories. Discriminative methods described in \cite{bertinetto2016learning,hariharan2016low} have the ability to update parameters of a base classifier that has learnt from training classes, while adapting to new classes for the specific task. But the complication in adapting classifiers in this manner is that they are prone to overfitting. Bertinetto \etal \cite{bertinetto2016learning} trained two-branch networks, where one branch receives an example and generates a set of dynamic parameters and second branch classifies the query image based on those parameters and a set of learnt static parameters. Noh \etal \cite{noh2016image} used a similar approach for question answering. Most of the existing works on one-shot learning focus on classification, not structured output. In \cite{caelles2017one}, a simple approach is proposed by authors to perform one-shot semantic segmentation by fine-tuning a pre-trained segmentation network on a labelled image. But, this method is prone to overfitting. Later, Shaban \etal \cite{shaban2017one} introduced a two-branched network to support dense semantic image segmentation in the one-shot setting. A N-way (classes) few shot segmentation framework based on metric learning and prototype learning has been introduced by Dong \etal \cite{dong2018few}. Very recently, Zhang \etal \cite{Zhang2018} proposed a one-shot semantic semantic framework with a new mask average pooling operation\cite{Zhang2018}. These earlier frameworks do not consider multi-scale information, and the performance is limited in real-world logo datasets.  

In contrast, instead of overly complicated network design, we use off-the-shelf neural network components in our architecture design that is trainable end-to-end. The contribution of the work lies in following aspects in terms of network design choices: $(1)$ We use a parameterized $1\times1$ convolution layer in order to measure the similarity in high-dimensional space. $(2)$ A multi-scale conditioning operation is used in order to handle the logos of different scale. $(3)$ Skip-connection between encoder-decoder parts of the network is found to be helpful for better information passing.

\begin{figure*}[t]
\label{pic2}
\includegraphics[width=1.0\linewidth]{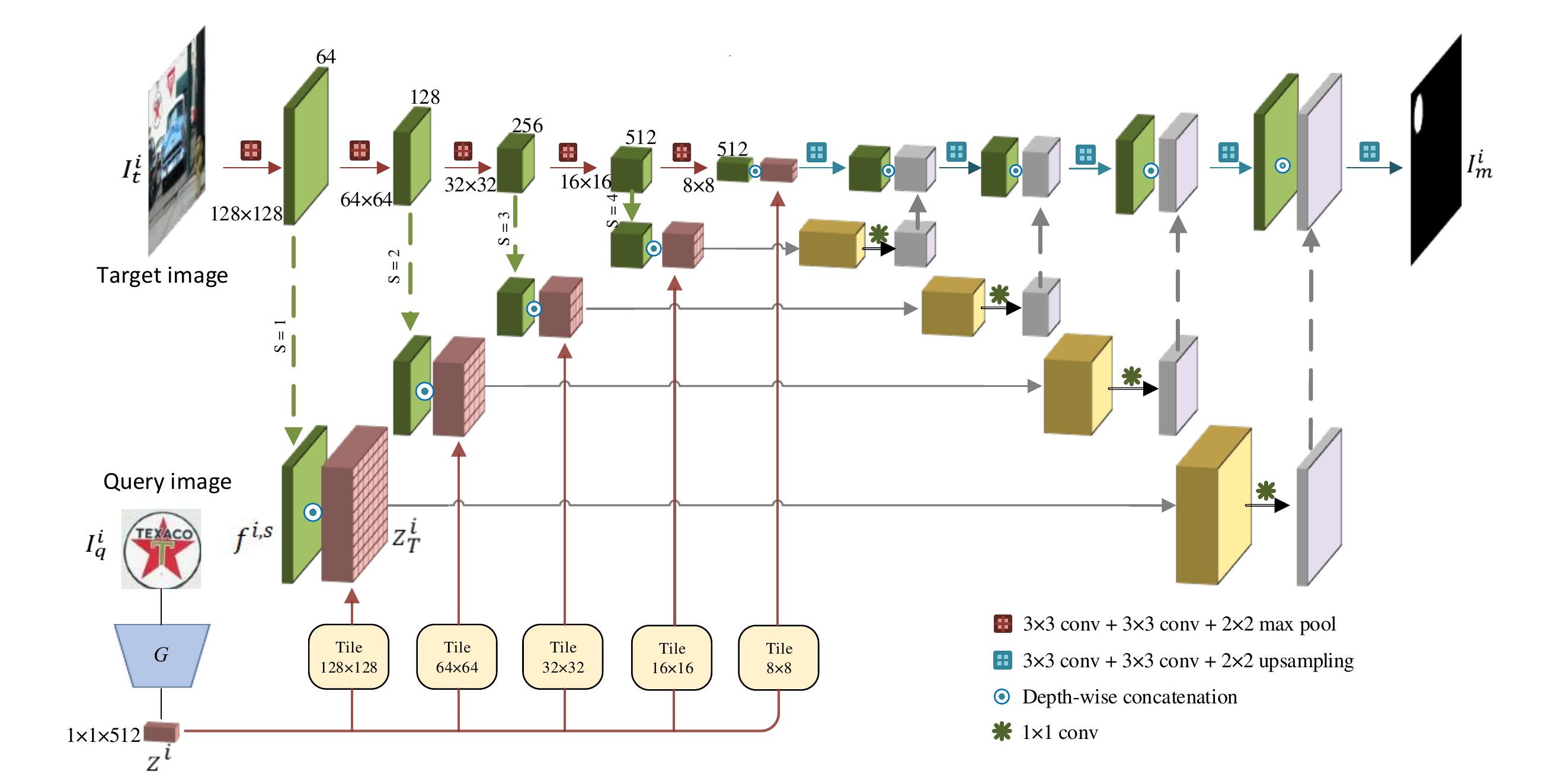}
\caption{Query-based logo detection framework: At-first, the conditional network $G$ takes the query logo image $I_{q}^{i}$ and outputs a latent representation $z_{i}\in \mathbb{R}^{{1}\times {1}\times 512}$. In the main segmentation branch, the target image $I_{t}^{i}$ is taken as input. After 5 stages of down-sampling, the encoder part reduces the input to $8\times8\times256$. The earlier obtained $z_{i}$ is concatenated with last layer of each stage via a tile operation. Suppose, at stage-1, i.e. s=1, the feature map $f^{i,s}$ is combined with the $z_{i}$ after the necessary tile operation. Then, after applying a $1\times1$ convolution on the combined representation, it is further concatenated with its respective layer at the decoder side as shown in the figure.   
   }
   
\label{fig:long}
\label{fig:onecol}
\end{figure*}

\section{Problem Setup}
The main objective of the logo detection problem is to find out whether a query logo of a particular company or organization is present in a target image or not. If it is present, we obtain a segmentation map containing information about the spatial location of the query in that target image. More formally, given a query image $I_{q}$ and a target image $I_{t}$ our job is to find out a segmentation map $I_{m}$ which is a binary 2D-matrix. The value $1$ in the segmentation map represents the region containing the query logo whereas $0$ represents background. During training, we have access to a large number of query target-mask triplets $\left \{{I_q^i,I_t^i,I_m^{i}}\right \}_{i=1}^N $ where $I_m^{i}$ is the semantic segmentation map of $I_t^i$ given the conditional query image $I_q^i$ of various logo classes. In the experiments, we have seen that our model is also able to generalize to unknown logos.
 
\section{Proposed Framework}
\subsection{Overview}
Our proposed end-to-end network can be divided into two steps: (i) conditioning step and (ii) segmentation step. In the conditioning part, we extract a conditional latent representation of the query image through a Convolutional Neural Network or CNN. The segmentation part of our model is a modified version of the U-Net \cite{ronneberger2015u} architecture. The encoder intends to extract more robust feature information from the target image and subsequently, the decoder tries to obtain the corresponding segmentation map conditioned on the latent representation of the query logo-image. 
By the word \textit{conditioning}, we here intend to segment regions of the target image conditioned on the query latent vector; in other words, the regions in target image which have a close similarity with the conditional query logo image. In order to overcome the possibility of drastic mismatch in scale and resolution between the query logo and the target image containing that logo, we encourage the model to learn \textit{multi-scale conditioning} by combining the logo representation at different scales to the encoder part of the U-Net architecture. The proposed one-shot learning framework is fully differentiable and the two-branch architecture can be trained in an end-to-end manner. 


\subsection{Conditioning Module}
Logo-to-dense feature map conversion is a primary step in our architecture. 
Examining the power of extracting robust task-specific feature representation of convolutional networks, we use a CNN as the feature extractor. For this purpose, at first, we resize each query logo image to a fixed size of $64 \times 64$. Next, we feed the query image $I_{q}^{i}$ to the network $G(.)$ which encodes the image to a latent representation $z^{i}$. Specially, the logo image is converted into a multichannel feature vector of unit spatial dimension (i.e. $1\times1\times512$) for further use. Thus,

\begin{equation}
z^{i} = G(I_{q}^{i};\theta_{G})
\end{equation}

where $z^{i} \in  \mathbb{R}^{1 \times 1 \times 512}$ and $\theta_{G}$ is the parameter of the network. We adapt VGG-16 like architecture consisting of 13 convolution layers with rectified non-linearity (ReLU) activation after each layer. 
After 6 max-pooling layers, the input image gets converted to $\frac{1}{2^6}$ shape of feature map giving a final feature representation of dimension $1\times1\times512$. 

\subsection{Multi-scale Segmentation module}
\label{multi_seg}

In this section, we describe our multi-scale segmentation module. Given a target image $I_{t}^{i}$ and the obtained latent representation $z^{i}$ of query logo, the segmentation network $F(.)$ will try to obtain a segmentation map $\overline{I_{m}^{i}}$. Mathematically,
\begin{equation}
\overline{I_{m}^{i}} = F(I_{t}^{i}, z^{i};\theta_{F})
\end{equation}
where $\theta_{F}$ is the parameters of the network. Following the architecture of U-Net, we have introduced two networks: Encoder network and Decoder network.  Encoder network encodes the target image to a latent representation. The encoder follows the typical structure of a CNN. It consists of a set of two convolution layers of $3\times3$ filter size and followed by a $2\times2$ max-pooling layer with stride 2. This set of operations are repeated five times to progressively down-sample the input image until we get a bottleneck layer where we get the final latent feature representation. In this five-stage down-sampling process we have used increasing number of filters: 64, 128, 256, 512 and 512 respectively in each stage. After, each stage of down-sampling the spatial dimension of the feature maps are reduced $\frac{1}{2}$ times. Thus, the final latent feature map is $\frac{1}{2^{5}}$ of the input image. For example, if $I_{t}$ has the shape of $256\times256\times3$, then the latent feature map is of $8\times8\times512$. We have seen that further down-sampling the obtained spatial representation eventually degrade the results to some extent. 


In order to learn the correlation between the query image and the target image, a naive approach would be to just concatenate the latent representations of the two networks.
However, the said method has several disadvantages: 1. The scale difference between the query and the target image makes it difficult to achieve robust segmentation with a single scale conditioning approach. Thus, multi-scale aggregation is required to obtain detailed parsing maps. 2. Even state-of-the-art logo detection frameworks find it hard to detect smaller logos and ones present in very large scenes. Recent works on object detection \cite{bell2016inside,hariharan2015hypercolumns,kong2016hypernet, wang2018multi} reveal that the feature map of the shallower convolution layers have higher resolution and are helpful to detect small objects whereas the deeper layers contain richer task-specific semantic information. 

\paragraph{Multi-scale Conditioning:} Taking cues from the aforesaid observations, we make use of the in-network hierarchical feature produced from feature maps of encoder part of the segmentation network having different spatial resolution. Let, given a target image $I_{t}^{i} \in \mathbb{R}^{H\times W\times C}$, the feature map extracted from the $s^{th}$ stage of the five-stage convolution network be $f^{i,s} \in \mathbb{R}^{H_{s}\times W_{s}\times C_{s}}$ , where, $H_{s}= \frac{H}{2^{s}}$ and ${W_{s}}= \frac{W}{2^{s}}$. Earlier, we got $z^{i} \in \mathbb{R}^{{1}\times {1}\times 512}$ as the conditional query logo representation. In order to impose condition on the segmentation network using latent query vector $z^{i}$, we use a technique similar to sliding window based template matching. To measure the similarity between latent query vector $z^{i}$ and each discrete feature $f_{mn}^{i, s}$ in $f^{i, s}$, we use a simple concatenation operation followed by a $1\times1$ convolution. 

\begin{equation}\label{conv_eqn}
\overline{f}_{mn}^{i,s} = Conv_{1\times1}^{s}([f_{mn}^{i,s}; z^{i}])
\end{equation}
where$f_{mn}^{i, s}$ represents the feature representation at position $(m,n)$, 
and which, upon concatenation with $z^{i}$ gives a vector of size $1\times 1\times (C_{s}+512$). The number of $1\times1$ convolutional filters in $s^{th}$ layer is set to $C_{s}$, and thus it gives $\overline{f}_{mn}^{i,s}$ of dimension $1\times 1\times C_{s}$. By this simple operation(eqn. \ref{conv_eqn}), it tries to capture the similarity between the latent query vector and feature embedding at $(m,n)$ position of the feature map $f^{i,s}$ from segmentation network. High activation values in $\overline{f}^{i,s}$ signifies high extent of matching that the query logo is likely to be present at that location and vice-versa. Mathematically, let’s denote conditional query representation as $Z$ and one discrete feature $f_{mn}^{i, s}$ as $F$. Using $1\times1$ convolution operation with parameters $\theta_{1\times1}$, we intend to model $P(Y| F, Z; \theta_{1\times1})$ where $Y$ denotes the similarity between $F$ and $Z$.  To illustrate,  $P(Y| F_{1}, Z_{1}; \theta_{1\times1}) > P(Y| F_{2}, Z_{2}; \theta_{1\times1})$ when the similarity between $F_{1}$ and $Z_{1}$ is higher than $F_{2}$ and $Z_{2}$. These parameters of $1\times1$ convolutions are learnt implicitly from the traditional cross-entropy loss in an end-to-end manner. In other words, this $1\times1$ convolution is a function that takes $Z$ and $F$ as input, and predicts their similarity at its output in terms of activation values. Unlike other works \cite{Zhang2018}, instead of using cosine-similarity computation for feature matching, we make such choice mainly because of two reasons: : \textit{First}, using a parameterized layer to capture the similarity is expected to be robust to various kind of deformations, illuminations present in real word logo retrieval scenarios. \textit{Second}, $1\times1$ convolution handles the intra-class variance well, in other words, sometimes it is found that despite same logo-class it has some difference in terms of the color of the logo, font used, presence of text inside a logo (for example in \textit{adidas} logos, sometimes the text remains missing). Also, it is to be noted that we use the final latent query representation from conditioning module for multi-scale conditioning at every layer in encoder part (segmentation network) since deeper latent representation is supposed to contain better semantic information of the query logo, instead of using pooled feature representation from earlier layers of conditioning module. More comparison with alternative choices are given in ablation study (Section \ref{ablation}).

For faster and efficient implementation, a tile operation is performed on $z^{i}$ to convert it to a specific spatial dimension such that is compatible for concatenation with $f^{i,s}$: $z_{T}^{i,s}=Tile(z^{i};s_{c})$, where $s_{c}$ is the scale required to tile $z^{i}\in \mathbb{R}^{{1}\times {1}\times 512}$ to a new dimension $z_{T}^{i,s} \in \mathbb{R}^{H_{s}\times W_{s}\times 512}$ and concatented with $f^{i,s}$ giving a dimension of  $\mathbb{R}^{H_{s}\times W_{s}\times 512 + C_{s}}$. This is followed by a $1\times1$ convolution with $C_{s}$ filters:
\begin{equation}\label{conv_eqn2}
\overline{F}^{i,s} = Conv_{1\times1}^{s}([f^{i,s}; z_{T}^{i,s}])
\end{equation}
where $\overline{F}^{i,s} \in \mathbb{R}^{H_{s}\times W_{s}\times C_{s}}$. Alternatively, it can be interpreted as applying a sliding window based template matching over the target image with stride $2^{s}$ (since every max-pooling layers steps down the spatial resolution of feature-map by a factor of $2$) for $s^{th}$ stage in encoder part of segmentation network.

The decoder network of the segmentation model is similar to the encoder part, the only difference being that up-convolution has been used. Every stage in the  decoder  path  consists  of  an  up-sampling layer followed by a $2\times2$ up-convolution layer. After that, we combine this with the previously obtained fused representation of the corresponding stage. At last, two $3 \times 3$ convolutions, followed by a ReLU completes the set of operations for a particular stage. These set of operations are repeated for all the stages until we get the predicted segmentation map $\overline{I_{m}^{i}}$ through a SoftMax layer.
At the end, we define a pixel-wise binary cross entropy loss $L$ which is used to train the complete model. 
\begin{equation}
L= \frac{1}{HW}\sum_{i=0}^{W} \sum_{j=0}^{H} (-I_{m}^{i}log(\overline{I_{m}^{i}})) 
 \end{equation}
The aim of the model is to minimize this loss by updating the parameter of the both the conditioning ($\theta_{G}$) and segmentation ($\theta_{F}$) modules of the network through back-propagation technique in an end-to-end manner. 

Some earlier works like Mask-RCNN \cite{he2017mask} first predicts the bounding box using Region Proposal Network(RPN) and then performs the instance segmentation as a two steps process. This requires two separate loss functions for region proposal and final instance segmentation, respectively. In contrast, we avoid using a two steps process and predict the segmentation map in a single step directly, and thus it makes the process faster as well as easy for implementation. That becomes feasible because of \textit{multiscale conditioning} and \textit{parameterized similarity matching layer} with \textit{skip connections} that helps in better information passing between encoder and decoder part of the framework.

\section{Experiment}

We have evaluated the performance of our framework in both one-shot and traditional settings. In the one-shot setting, given an unseen conditional query logo image (from the test set), the model trained on a completely disjoint set of logo classes returns a reference binary mask for localization. In the traditional setting, on top of our query-based strategy, we experiment with the traditional train-test split strategy used by previous works \cite{su2017deep} where some images for each logo-class are utilized for training while the rest is used for evaluation. Despite our one-shot architecture, we evaluated the framework in the traditional setup in order to directly measure our model's performance against the traditional setting that has been adopted by every existing work. Although different classification and recognition problems have been extended to query-based retrieval setups (e.g. image classification to image retrieval \cite{Gordo2016}, handwriting recognition to query-based word spotting \cite{Manmatha1996}), not many prior works using deep learning frameworks exist that addresses logo retrieval in an open-vocabulary scenario and is robust to challenges like low resolution, cluttered background etc. While existing methods are designed for more complicated tasks (e.g. detection and localization of all the logos in the training set), the boost in performance even in sparse training data plus the ability to generalize to unseen logos in the wild justifies our proposal to extend the problem-statement from a traditional to query-based setup.

\subsection{Datasets}

Our model makes use of binary masks as ground truths. FlickrLogos-32 \cite{RombergICMR2011} is a very popular logo dataset consisting of boundary box annotations as well as binary masks. Therefore, we train our model on this dataset. To evaluate the robustness of the one-shot architecture we explored another dataset Toplogos-10 \cite{su2017deep}.  The main reasons to select FlickrsLogos for training and TopLogos for testing are as follows: (a) To the best of our knowledge, FlickrsLogos is the only logo dataset which has binary segmentation mask as ground truth along with bounding box labels. Rest of the available datasets have bounding box labels only. (b) During testing, we preferred to use TopLogos because it has instances of logos of varying sizes and different cluttered background scenario which provides many challenges for logo detection. Brief discussions of these datasets are given below.

\textbf{Training Set} - \textit{FlickrsLogos-32:}
This dataset comprises 8,240 images from 32 different logo classes, each class representing a particular brand. Each class has 70 images with ground truth annotations in the form of bounding boxes and binary masks. 
To make the dataset congenial for our approach, we ignore 6,000 images with no logo class. Thus, we have total $32\times70$ i.e. 2240 images available for our experiments.

\textbf{Evaluation set} - \textit{TopLogos-10:}
It contains 700 images of 10 different clothing brand logos with various degree of composition complexity in the logos. Basically, there are ten logo classes: “Adidas”, “Chanel”, “Gucci”, “Helly Hansen”, “Lacoste”, “Michael Kars”, “Nike”, “Prade”, “Puma”, “Supreme”. For every class, there are 70 images with fully manually labelled logo bounding boxes. But, unlike the FlickrsLogos-32 it does not have binary mask annotations. 
This dataset contains natural images where logo instances in a variety of context, e.g. Hats, wallets, shoes, shown gels, lipsticks, spray, phone covers, eye glasses, jackets, peaked caps, T-shirts, and sign boards. In short, Toplogos-10 represents logo instances with varying sizes in natural real-world scenarios that provide real challenges for logo detection. The main reason behind using TopLogos-10 dataset is to check the generalization ability of our framework in a completely unseen scenario with more complexity, which is a pressing need of one-shot framework. Even if TopLogos has only 10 logo classes, every query-target pair (note that, this number is large) is unknown to the trained model during testing and it challenges the model to deal with real one-shot scenarios like varying logo sizes, different cluttered background etc.

\subsection{Implementation Details}
We will discuss some salient details of our model: starting from data preparation to training and inference details. 

\textbf{Data preparation: } \textit{(a) One-shot setting: }As discussed earlier, we have used FlickrsLogos-32 dataset for training our model in one-shot setting. It contains 2240 images with ground truth annotations. We denote this set as $D^{t}$ and their corresponding masks as $D^{m}$. The images of these two sets are resized to a fixed dimension $256\times256$. Now, we obtain the conditional query logo objects by cropping the main images with respect to their bounding boxes followed by resizing to a dimension of $64\times64$. This forms the conditional set of query objects $D^{q}$. Then, we generate a large number of triplets from the obtained image sets $\lbrace D^{q}, D^{t}, D^{m}\rbrace$. If $n$ is the number of images per logo class then $n\times(n-1)$  triplet combination is possible per class. In total, for 32 logo classes we get $32\times(70\times69)$ i.e. 154560 triplets. We use 90\% of these triplets as our training set and 10\% triplets as validation set.
\textit{(b) Traditional setting: }In this set-up, our main objective of query-based logo search remains unchanged. But, now we generate our training triplets only from a small set (40 images per class) of the available data in the FlickrsLogos-32 dataset. These triplets will be used for training the model. And for testing we will use the rest of the images for each class.

\textbf{Training: } We train our model in an end-to-end fashion from scratch using these large number of generated triplets $\left \{{I_q^i,I_t^i,I_m^{i}}\right \}_{i=1}^N $ where $N$ is the total number of triplets. The conditioning branch takes $I_q^i \in \mathbb{R}^{64\times64\times3}$ as an input and obtains a conditional representation of the logo object. This representation will be concatenated with the segmentation network at different scale. The segmentation branch takes $I_t^i \in \mathbb{R}^{256\times256\times3}$ as input and tries to predict a binary segmentation mask $I_m^i$. We use Gaussian initialization with 0.01 standard deviation. We train our model using stochastic gradient descent (SGD) optimizer with initial learning rate 0.0004 and momentum and weight decay are set to 0.9 and 0.0005 respectively. The model is trained for 100K iterations with batch size of 32 for optimizing loss up to a satisfactory level. We have implemented the whole model in Tensorflow and run on a server with Nvidia Titan X GPU with 12 GB of memory.

\textbf{Testing: } In one-shot setting, the main advantage of our model is that once it's trained it can be used for any logo class. To test the scalability and robustness of our model we have used a different dataset. For traditional setting, we test our model on the same FlickrsLogos-32 dataset but with testing images available for each class.

The final binary segmentation mask is obtained by using a threshold of 0.5 on the predicted mask.  To obtain quantitative results on these datasets, we generate a minimum bounding box for each logo instances covering the mask from the predicted binary mask. To attain bounding boxes from the output map, first, we compute the topmost spatial position $T$ ($x_{t}$ , $y_{t}$), bottom most binary pixel position $B$ ($x_{b}$ , $y_{b}$), leftmost binary pixel position $L$ ($x_{l}$, $y_{l}$) and topmost binary pixel position  $R$ ($x_{r}$ , $y_{r}$). Then, we evaluate bounding box value ${X}'=x_{l} $, ${Y}'=y_{t} $, ${H}'=y_{b}-y_{t}$ and ${W}'=x_{r}-x_{l}$.
At last, these bounding boxes are compared with the ground-truth bounding boxes to obtain the Intersection over Union (IoU). 
For the performance evaluation, we have used mean Average Precision (mAP) metric for all classes. For the mAP calculation, we choose the  IoU threshold as 0.5. It means that a detection will be considered as positive if the IoU between the predicted and ground-truth exceeds 50\%. As the proposed logo detection technique is based on segmentation of the logo, we have also included the mean pixel IoU or mPixIoU \cite{liang2015semantic} as another metric for the evaluation purpose. However, unlike FlickrsLogos-32 dataset, other dataset does not have any pixel level binary annotation. Thus, the mPixIoU metric is shown only for FlickrsLogos-32 dataset. The code is available here \footnote{https://github.com/AyanKumarBhunia/Deep-One-Shot-Logo-Retrieval.}.

\subsection{Baselines}
To exploit the robustness of our approach, we compared the following baselines with our proposed approach.

\textbf {Fine-tuning: }As suggested by \cite{caelles2017one}, we fine-tune a pre-trained FCN network (only the fully connected layers) with full supervision on the available paired data of a particular logo class and test it on the target images.

\textbf{U-Net:} Here, we focus on the naive way for generating segmentation maps by considering each logo class as a separate semantic class. We use U-Net \cite{ronneberger2015u} encoder-decoder network that takes the target image as input and outputs segmentation map without any conditional reference. Note that this set-up cannot be extended to new classes. 

\textbf{SiameseFCN:} Siamese networks are extensively used for one-shot classification tasks \cite{koch2015siamese}. In this method, we use two pre-trained FCNs to extract dense features from query images and target images. L1 similarity metric is used to learn the coherence between the feature for every pixel in the query image and the target image in order to produce a pixel-level binary mask.


\textbf{CoFCN:}
Here, we explore a conditional segmentation network based on FCN \cite{shaban2017one}. At first, we feed query images in the conditional branch to generate a set of parameters $\theta$. We use $\theta$ in the parameterized part of the learned segmentation model, which takes target images as input and produces a segmentation map.

\textbf{SG-One:} This one-shot segmentation framework \cite{Zhang2018} uses a new masked average pooling operation to extract the latent representation of the query image. However, since our framework uses cropped query image (i.e. support set), we use simple global average pooling instead, and use cosine-similarity followed by a tanh layer for similarity matching with rest setup similar to \cite{Zhang2018}. A quite similar one-shot framework has also been addressed for visual tracking problem \cite{Zhang2018}.  

\begin{table}[]
\centering
\caption{Comparison of Logo detection performance on FlickrsLogos-32 dataset following traditional setting.``\#'' denotes query-based frameworks.}
\begin{tabular}{l|ccc}
\hline
\textbf{Method}                        & \textbf{mPixIoU} & \textbf{mIoU} & \textbf{mAP} \\ \hline
Bag of Words (BoW) \cite{romberg2013bundle}                     & -                & -             & 54.5         \\ \hline
Deep Logo \cite{iandola2015deeplogo}                               & -                & -             & 74.4         \\ \hline
BD-FRCN-M \cite{oliveira2016automatic} & -                & -             & 73.5         \\ \hline
RealImg \cite{su2017deep}                               & -                & -             & 81.1         \\ \hline
Faster-RCNN \cite{ren2015faster}                               & -                & -             & 70.2         \\ \hline
SSD \cite{liu2016ssd}                               & -                & -             & 67.5         \\ \hline
YOLO \cite{redmon2016you}                               & -                & -             & 68.7         \\ \hline
U-Net \cite{ronneberger2015u}                                  & 20.3             & 30.1          & 40.5         \\ \hline
\textsuperscript{\#}SiameseFCN \cite{koch2015siamese}          & 70.8             & 77.6          & 79.4         \\ \hline
\textsuperscript{\#}CoFCN \cite{shaban2017one}                 & 71.4             & 81.9          & 84.1         \\ \hline
\textsuperscript{\#}SG-One \cite{Zhang2018}                 & 76.5             & 84.5          & 86.9         \\ \hline
\textsuperscript{\#}Ours              & \textbf{78.2}             & \textbf{86.7}          & \textbf{89.2}         \\ \hline
\end{tabular}
\label{tab1}
\end{table}

\begin{table}
\centering
\caption{Logo detection performance on Evaluation set following one-shot setting}
\label{tab2}
\begin{tabular}{c|c|l|c|c|c} 
\hline
 \textbf{Training set}                                                                         & \textbf{Testing set}                                                                           & \textbf{Method} & \textbf{mPixIoU} & \textbf{mIoU} & \multicolumn{1}{l}{\textbf{mAP } }  \\ 
\hline
\multirow{5}{*}{\rotcell{\begin{tabular}[c]{@{}c@{}}FlickrsLogos\\(20 classes) \end{tabular}}} & \multirow{5}{*}{\rotcell{\begin{tabular}[c]{@{}c@{}}FlickrsLogos\\(12 classes) \end{tabular}}} & Fine-tuning     & 15.8             & 24.1          & 27.4                                \\ 
\cline{3-6}
                                                                                               &                                                                                                & SiameseFCN      & 36.9             & 46.5          & 51.1                                \\ 
\cline{3-6}
                                                                                               &                                                                                                & CoFCN           & 44.6             & 51.3          & 59.7                                \\ 
\cline{3-6}
                                                                                               &                                                                                                & SG-One          & 52.3             & 59.7          & 64.4                                \\ 
\cline{3-6}
                                                                                               &                                                                                                & \textbf{Ours}   & \textbf{59.1}    & \textbf{66.3} & \textbf{66.8}                       \\ 
\hline
\multirow{5}{*}{\rotcell{\begin{tabular}[c]{@{}c@{}} FlickrsLogos\\(20 classes)\end{tabular}}} & \multirow{5}{*}{\rotcell{\begin{tabular}[c]{@{}c@{}} TopLogos\\(10 classes)\end{tabular}}}     & Fine-tuning     & -                & -             & 14.7                                \\ 
\cline{3-6}
                                                                                               &                                                                                                & SiameseFCN      & -                & -             & 24.4                                \\ 
\cline{3-6}
                                                                                               &                                                                                                & CoFCN           & -                & -             & 29.2                                \\ 
\cline{3-6}
                                                                                               &                                                                                                & SG-One          & -                & -             & 36.9                                \\ 
\cline{3-6}
                                                                                               &                                                                                                & \textbf{Ours}   & -                & -             & \textbf{40.1}                       \\
\hline
\end{tabular}
\end{table}

\begin{figure*}
\label{pic3}
\begin{center}
  \includegraphics[width=1\linewidth]{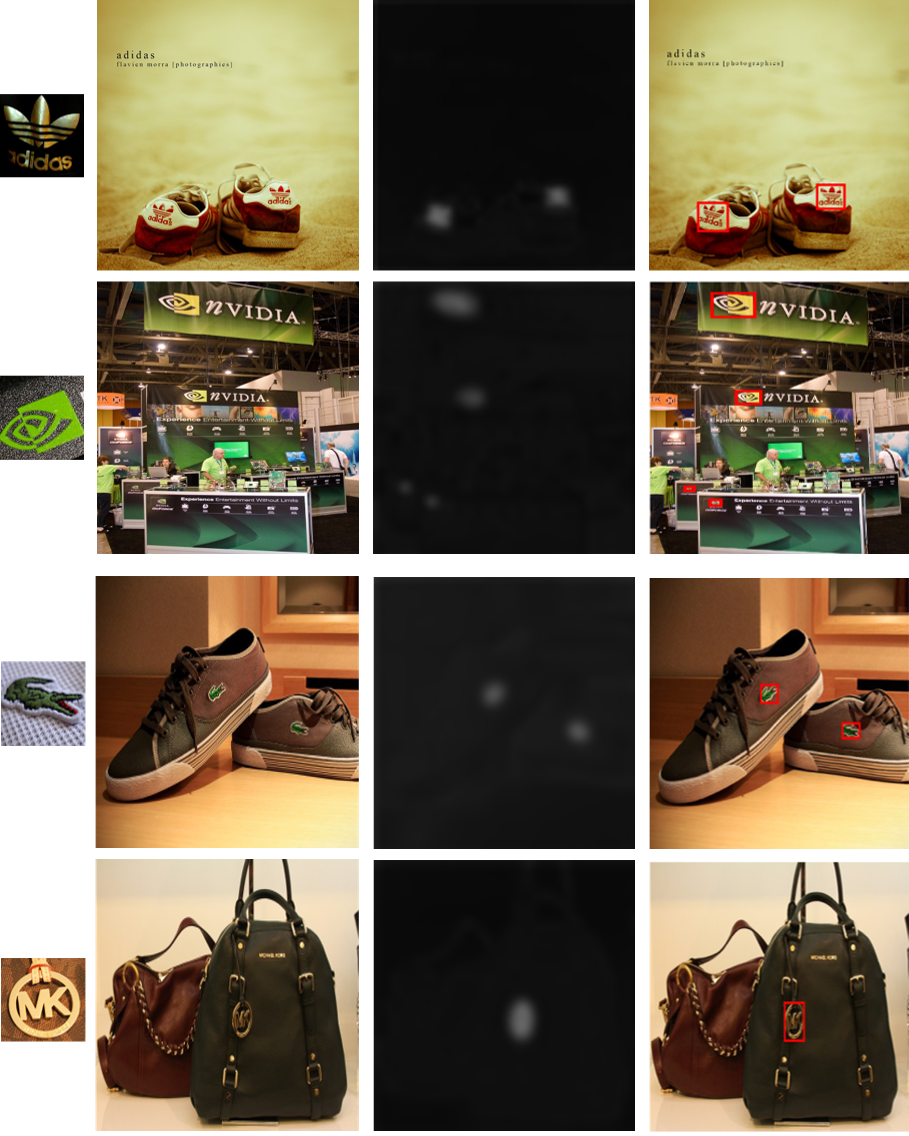}
  \caption{Some qualitative results of our method: In the first two rows we have shown results on FlickrLogo-32 using traditional setting and for the next two rows we have shown one-shot results on TopLogos-10 dataset. I Here, images are in the following order (left to right): query logo, target image, predicted mask, and final detection result (shown by boxes on the images),respectively.}
\end{center}
\end{figure*}

\subsection{Evaluations}

\begin{figure}
  \includegraphics[width=1\linewidth]{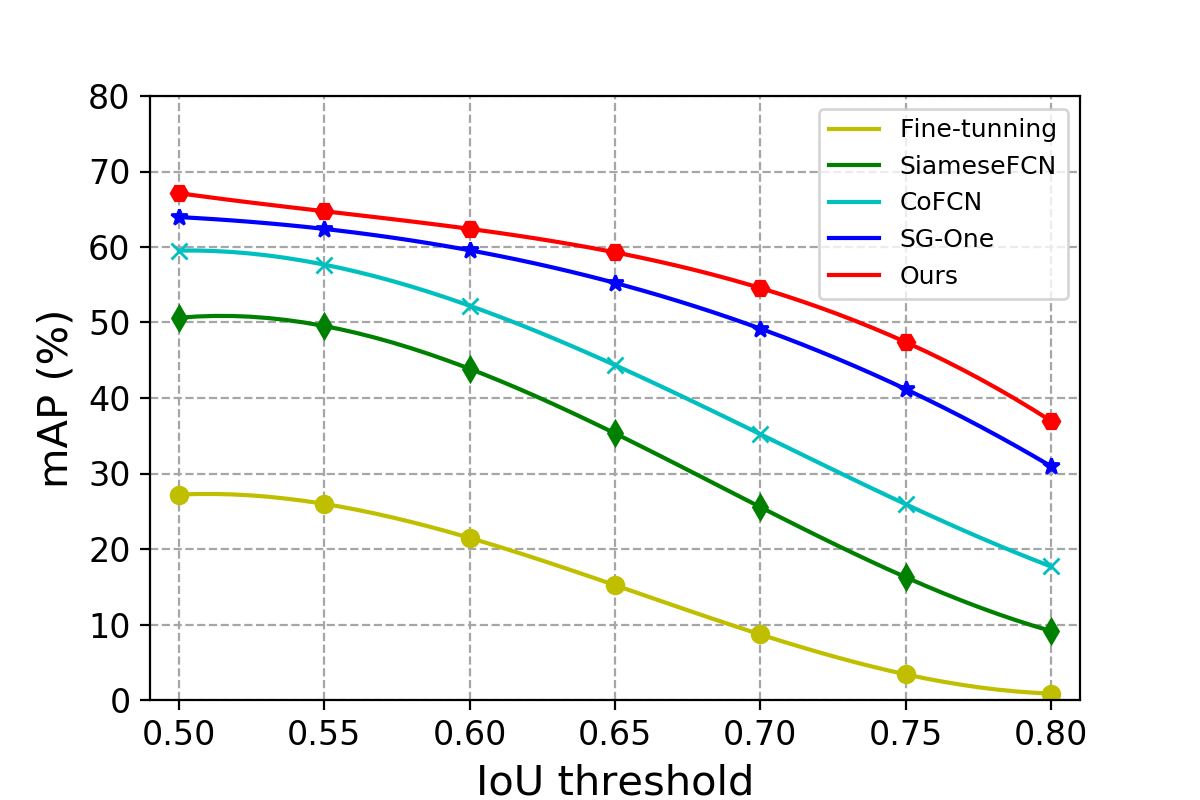}
  \caption{Performance with varying IoU thresold values.}
  \label{figure4}

\end{figure}

\textbf{Traditional setting:}
We have compared our results with the baselines and some state-of-the-art logo detection methods \cite{iandola2015deeplogo, oliveira2016automatic,romberg2013bundle,su2017deep} as well as evaluated the performance using popular bounding box detectors like Faster-RCNN\footnote[1]{https://goo.gl/zQsSka (Faster-RCNN)} \cite{ren2015faster}, YOLO\footnote[2]{https://goo.gl/vUFEuq (YOLO)}\cite{redmon2016you} and SSD\footnote[3]{https://goo.gl/vxR9c7 (SSD)} \cite{liu2016ssd} while opting for VGG as the baseline architecture. Due to limited data, we initialize this framework with weights trained on Pascal-VOC dataset (see Table \ref{tab1}). Among popular traditional logo detectors, RealImg \cite{su2017deep} gives highest mAP score of 81.1\% and uses a deep model to synthetically generate training data in order to improve its performance. Though state-of-the-art object detectors like Faster-RCNN, YOLO, SSD achieve impressive performance on benchmark object detection datasets, their performance is limited in FlickrLogos-32 dataset because of limited training data. For the same reason, the performance of U-Net is poor.
SiameseFCN, which is a query-based framework, performed moderately on FlickrLogo-32 dataset. CoFCN performed well with a high mAP of 84.1\% which is 3.0\% and 9.4\% greater than RealImg \cite{su2017deep} and Deep Logo \cite{iandola2015deeplogo}  respectively. The proposed multi-scale query based detection technique achieved a better mAP than the previous state-of-the-art method \cite{su2017deep} with 8.1\% increase in absolute mAP; our rival query-based method, SG-One \cite{Zhang2018} trails by 2.3\% mAP value. 
\\
\textbf{One-shot setting:} 
For the one-shot setup, the results shown in Table \ref{tab2} indicate the capability of different methods to generalize to new classes. These results are comparatively lower than the results observed in the traditional setting. We realize that this is because we are trying to evaluate the performance of the model on a more open space setting by allowing the system to detect logo in a target scene from a single reference logo sample; contrarily, in the traditional setup, the total number of logo classes is fixed and 40 images from each logo class were used to train the model. The fine-tuned baseline produces relatively low mAP since it quickly overfits to the fine-tuning data in the support set. For FlickrLogo-32, our proposed method outperformed the general fine-tuning based approach by 39.4\% mAP value. On TopLogos-10, which is a clothing-logo dataset with dense background clutter, observed results are relatively low. We demonstrate some qualitative results of our framework in Figure 3. We have also evaluated the performance by varying IoU threshold from $0.5$ to $0.8$ and the performance of our framework drops to a limited extent compared to other competitive methods (see Figure \ref{figure4}), and thus it illustrates the superiority of our design choices.

These improvements over other existing one-shot frameworks \cite{koch2015siamese, shaban2017one, Zhang2018} are mainly due to following design choices in our framework. (1) Multi-scale conditioning (2) Parameterized similarity matching operation through
$1\times1$ convolutional layer at every scale (3) Skip connection between encoder and decoder part of the network. Overall, our framework is undoubtedly simple, easy to implement and understand.


\begin{table}[]
\label{tab3}
\centering
\caption{mAP (\%) of different variants of our configuration on FlickrsLogos-32 dataset using traditional setting}
\begin{tabular}{c|c|c|c|c}
\hline
\multirow{2}{*}{\textbf{Method}} & \multicolumn{3}{c|}{\textbf{k-shot setting}} & \multirow{2}{*}{\textbf{\begin{tabular}[c]{@{}c@{}}Traditional\\ Setting\end{tabular}}} \\ \cline{2-4}
 & k=1 & k=3 & k=5 &  \\ \hline
Variant 1 & - & - & - & 36.5 \\ \hline
Variant 2 & 59.1 & 64.5 & 65.6 & 82.7 \\ \hline
\textbf{Proposed} & \textbf{66.8} & \textbf{71.2} & \textbf{71.6} & \textbf{89.2} \\ \hline
\end{tabular}
\label{tab3}
\end{table}

\subsection{Ablation Study}\label{ablation}
In this section, we have given a comprehenssive analysis of each sub-variant of our framework along with quantitative analysis for different alternative network design choices. Here, we have also shown results for $k$-shot cases, using $k$ different samples of the query image as our conditional input serially, and final binary mask is obtained by logical OR operation between the $k$ predicted binary maps. For the traditional set-up we have used  FlickrsLogos-32 dataset as before. For $k$-shot setting, we have trained our model using images of 20 classes of FlickrsLogos-32 dataset and the remaining part is used for testing (Table \ref{tab3}).


\begin{itemize}[leftmargin=*]

\item {{Variant 1}: } Here, we use a general U-Net architecture without any conditioning branch(not a query-based framework) and notice limited performance. 


\item {{Variant 2}: }In this variant, we use conditional query representation, which is concatenated with the latent space of segmentation module, in order to fetch of the location of query logo in a target image. However, it fails to detect small logos in images, specifically when the size of the query logo image and the actual logo present in the target differs considerably. 


\item {{Proposed Method}: }We overcome the above problems by employing a \textit{multi-scale conditioning} operation. This makes our model capable of detecting small transformed logos, which is the one of the major challenges for logo detection.
\end{itemize}

\noindent\textbf{Alternative Designs:} To localize the query logo within the target scene image, instead of cosine-similarity measure, we focus on simple concatenation operation followed by $1\times1$ convolution to learn the correlation. Using a parameterized layer for similarity matching at multiple-scales offers a better performance in the context of logo detection (see section \ref{multi_seg}). Keeping rest of the architecture same, we calculate cosine similarity at each spatial position of feature-map from encoder (segmentation) network with conditional query latent vector (converted to the same depth as $f^{i,s}$ using $1\times1$ convolution), followed by a $tanh$ layer and finally combine to the corresponding decoder stage. This alternative network design achieves mAP values of 64.1\% and 87.8\%, trailing ours by 2.7\% and 1.4\% for one-shot and traditional setting on the FlickrsLogos-32 dataset (same setting as Table \ref{tab3}).
Also, we tried using pooled (Global Average Pooling \cite{lin2013network}) feature representation of feature maps from multiple layers of the conditional network, with an intuition to get a multi-scale representation of the query logo and evaluate similarity (\textit{ours}) with feature maps of same stage from segmentation network; the results, however, drop by 5.6\%. This signifies that the final latent representation from deep layers of the conditional network contains higher semantic information of the logo that can be used as a conditional input to segmentation network to fetch the logo at multiple-scales of the target image.

\vspace{0.5cm}

\begin{figure}
\centering
  \includegraphics[width=0.55\linewidth]{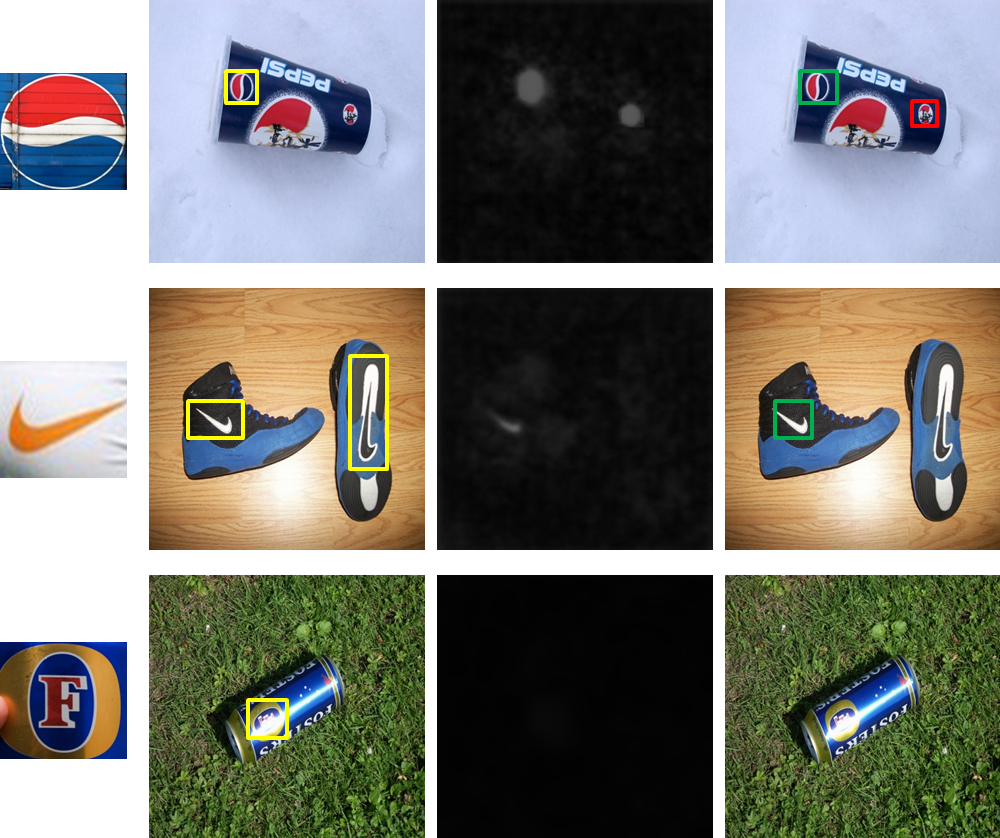}
  \caption{Some failure cases of our frameworks. Yellow box is the Ground-Truth. Green box is True Positive. Red box is False Positive. First row illustrates a false-positive case where our model wrongly predicts logo instances(marked in red). This happens due to certain extent of similarity between the query logo and irrelevant patch in the target image. In the second row, it fails to detect a logo instances due to different orientation. Third row shows a failure case due to tiny size of the logo and irregular illumination.  
}
  \label{figure5} 
\end{figure}

\noindent\textbf{Error Analysis:} Figure \ref{figure5} shows some of the failure cases of our framework. Here, we have used simple pixel-wise cross-entropy loss to train the network; however,  rotation invariant network design could be explored on the top of our proposed multi-scale conditioning framework, so that rotation related error (see second row of \ref{figure5}) could be avoided. Adversarial feature deformation \cite{kumar2019handwriting} could be used to generalize the model from limited data. In addition, it is to be noted that in spite of multiscale-conditioning it fails in few cases for tiny size of the logo. To alleviate such errors, a feature magnification module like \cite{li2017perceptual} or attention mechanism \cite{bhunia2019texture} could be helpful. In addition, the imbalance between  foreground-background could be better handled using newly proposed focal-loss \cite{lin2017focal} which adds a multiplying factor to the normal cross-entropy loss in order to emphasize on hard, misclassified examples.

\section{Conclusion}
In this paper, we have re-designed the traditional logo detection problem setting by proposing a query based logo search system that uses a one-shot architecture. The driving idea of our architecture is the use of \textit{multi-scale conditioning} with a skip-connection based architecture that predicts a logo segmentation mask. The proposed framework is simple and easy to implement. It is capable of detecting new logo classes without additional training data.

We demonstrate the effectiveness of our system by doing experiment on unseen logos in the wild. From the experiments on publicly available logo detection datasets, we noted that our proposed system outperformed the benchmark results \cite{iandola2015deeplogo, oliveira2016automatic,romberg2013bundle,su2017deep} without even extending existing logo datasets. 

Though our system is scale invariant, it may fail for logos which are tiny in size. A feature magnification module \cite{li2017perceptual} could be useful to improve the performance in such cases. Also, the imbalance between fore-ground and background information could be handled properly to boost the performance further. In future we plan to  work on these to improve the accuracy.




\bibliographystyle{elsarticle-num}
\bibliography{sample}







\end{document}